\documentclass[runningheads]{llncs}
\usepackage{amssymb}
\usepackage{multirow}
\usepackage{graphicx}
\usepackage{amsmath}
\usepackage{subcaption} 
\usepackage[hidelinks]{hyperref}

\usepackage{makecell}
\usepackage{multirow}
\usepackage{booktabs} 
\usepackage{caption}
\usepackage{cleveref}

\usepackage[table]{xcolor}
\newcommand{\myparagraph}[1]{\smallskip\noindent\textbf{#1}}

\begin{document}
%


\title{Hard Negative Sample Mining for Whole Slide Image Classification}

%
%

\author{Wentao Huang\thanks{Email: wenthuang@cs.stonybrook.edu.}\inst{1} \and
Xiaoling Hu\inst{2} \and
Shahira Abousamra\inst{1} \and 
Prateek Prasanna\inst{1} \and
Chao Chen\inst{1}
}
\authorrunning{W. Huang et al.}
%
\institute{Stony Brook University, Stony Brook, NY, USA \and
Harvard Medical School, Boston, MA, USA\\
}

\maketitle              
\begin{abstract}


Weakly supervised whole slide image (WSI) classification is challenging due to the lack of patch-level labels and high computational costs. 
State-of-the-art methods use self-supervised patch-wise feature representations for multiple instance learning (MIL).
Recently, methods have been proposed to fine-tune the feature representation on the downstream task using pseudo labeling, but mostly focusing on selecting high-quality positive patches. 
In this paper, we propose to mine hard negative samples during fine-tuning. 
This allows us to obtain better feature representations and reduce the training cost. Furthermore, we propose a novel patch-wise ranking loss in MIL to better exploit these hard negative samples. Experiments on two public datasets demonstrate the efficacy of these proposed ideas. Our codes are available at \url{https://github.com/winston52/HNM-WSI}.

\keywords{Whole Slide Image \and Self-Training \and Hard Sample Mining.}
\end{abstract}

\section{Introduction}

Histopathology image analysis serves as the gold standard for cancer diagnosis and treatment \cite{lu2021ai,niazi2019digital,barisoni2020digital}. 
Due to the large size of the WSI, the heterogeneity of the tumor microenvironment, and the absence of patch-level labels, Multiple Instance Learning (MIL) \cite{dietterich1997solving} schemes are often applied to perform a prediction at the whole slide level. 
In MIL, each slide is considered a bag. A slide is partitioned into patches to create the instances within the bag. One challenge is that only bag-level (i.e. slide-level) labels are available, but not instance-level labels. 
Specialized training algorithms have been proposed to learn to make instance-level predictions and aggregate them for bag-level prediction~\cite{shao2021transmil,ding2023multi,li2021dual}.

\myparagraph{Feature representation learning.}
The performance of MIL heavily relies on feature representation of instances (patches). Due to the huge image size, end-to-end learning is computationally infeasible. Earlier work used convolutional neural networks (CNNs) pretrained on ImageNet to generate patch features \cite{ilse2018attention}. These features are then used for downstream MIL. Recently, more advanced self-supervised learning techniques such as SimCLR \cite{chen2020simple} and DINO \cite{caron2021emerging} have been applied to pretrain features using public histopathology image datasets. These features are much more biologically relevant and deliver better performance when combined with MIL \cite{li2021dual,chen2022scaling,zhang2023precise,kapse2024attention}. 

\begin{figure}[t]
\centering
\includegraphics[width=0.8\textwidth]{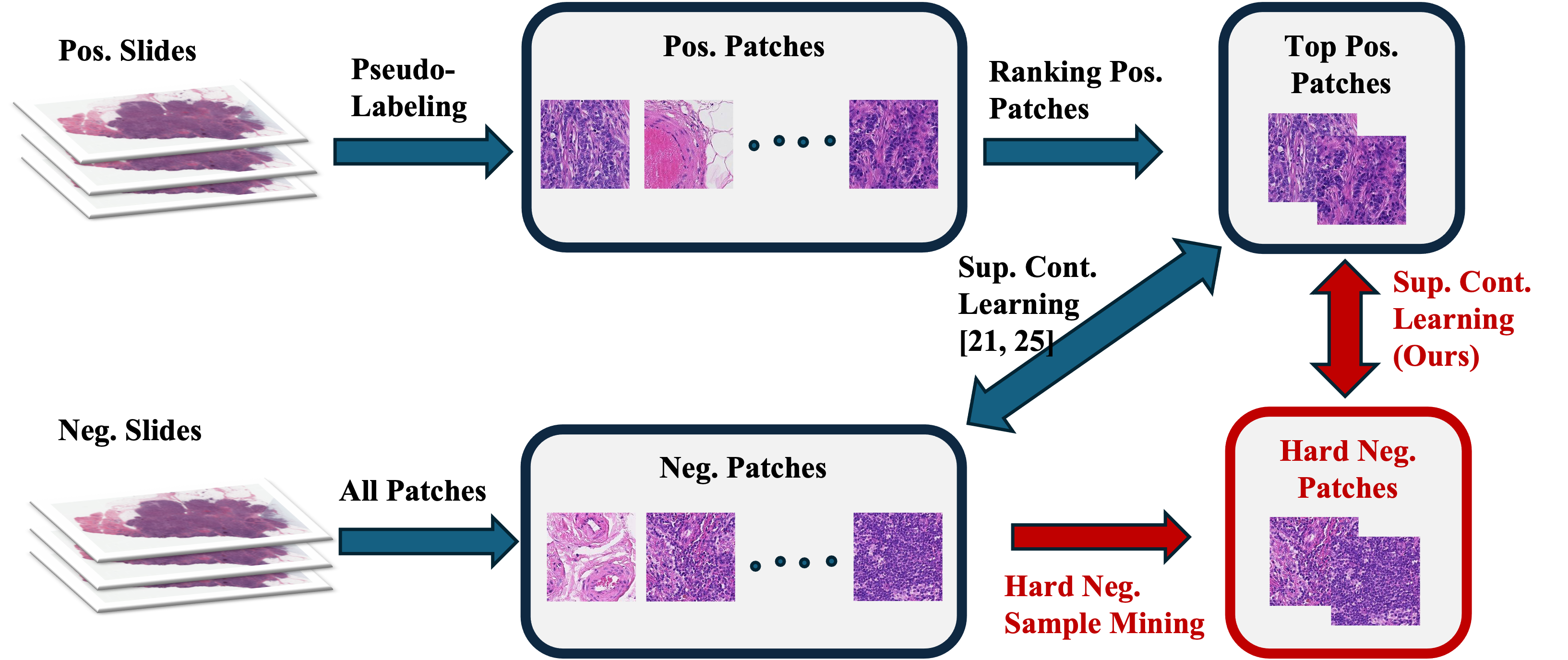}
\caption{Feature representation tuning. Previous methods \cite{liu2023multiple,qu2023rethinking} perform contrastive learning between top-ranked positive patches and all negative patches. Our method (highlighted in red) selects hard negatives for supervised contrastive learning.}
\label{fig:self-training}
\end{figure}

Despite the strong performance of these features, one cannot help but wonder whether they can be further tuned to better fit the downstream prediction task. To this end, new methods \cite{liu2023multiple,qu2023rethinking} have been proposed to use supervised contrastive learning to fine-tune the features. Patches are assigned pseudo-labels using a weak patch classifier from the downstream MIL. Supervised contrastive learning is carried out to ensure that patches of the same label are closer to each other in the feature space, and patches of different labels are far away from each other. However, these patch-level pseudo-labels can be noisy, and thus can derail the contrastive learning, leading to deteriorated features. To address this challenge, it was proposed to rank all patches with positive pseudo-labels based on model confidence, and select the top ones for learning. Meanwhile, since all negative slides only contain negative patches, we can just use these patches knowing that they are truly negative patches. 

\myparagraph{Our contribution: hard negative sampling for patch representation learning.}
In this paper, we question the design choice of these self-training methods regarding negative sample selection. While it is true that all patches from negative slides are true negative samples, they do not necessarily contribute to learning equally. In particular, we hypothesize that some negative patches are particularly useful in learning. To this end, we propose a novel hard sample mining algorithm to find negative patches that are particularly close to positive patches in feature representation. By focusing on these ``hard'' negative samples during contrastive learning, we achieve much better patch features for the downstream MIL. Moreover, since we only use a fraction of the negative instances, we are able to reduce the training time considerably. 
See \Cref{fig:self-training} for illustration. 

Indeed, the learning power of these hard negative samples can be further exploited in the downstream MIL. As a second technical contribution, we introduce a novel 
multiple instance ranking loss 
that pairwise compares the patch-level classifier's predictions on top positive samples and hard negative samples. By ensuring that the classifier ranks positive and negative patches correctly in terms of their ``positiveness'', we improve the instance-level classifier and thus the whole-slide-level prediction. 

In practice, we perform feature representation tuning and MIL training iteratively to achieve superior performance. Extensive experiments on two public datasets demonstrate the effectiveness of our proposed framework.

\section{Related Work}
\myparagraph{Multiple instance learning in WSI analysis.}
Multiple Instance Learning \cite{dietterich1997solving} (MIL) is a weakly supervised learning framework that can utilize coarse-grained bag labels to train a model when fine-grained instance annotations are not available. 
The MIL framework for WSI classification is divided into two groups: instance-based and bag-based. The instance-based method first predicts the probability of all instances and then aggregates these to obtain a bag prediction using Mean Pooling or Max Pooling \cite{campanella2019clinical,zhang2022dtfd}. In contrast, the bag-based method involves aggregating the embeddings of all instances into a single bag embedding and then classifying it using a bag classifier. Most current bag-based methods are Attention-based MIL \cite{ilse2018attention,li2021dual,qu2022bi} methods and ViT-based MIL \cite{shao2021transmil,ding2023multi} methods. Various strategies have been proposed to find positive patches more accurately \cite{lu2021data,li2021dual,tang2023multiple}. 
In this paper, we mainly develop an effective and efficient method by focusing on hard negative patches to improve the performance of the WSI classification.

\myparagraph{Self-training and pseudo labeling in WSI analysis.} Self-training is a widely used technique in semi-supervised learning \cite{lee2013pseudo,wei2021crest,xu2023toposemiseg}. The key idea is to generate pseudo-labels of unlabeled data using a model trained with labeled data and then train the model based on the combination of the labeled data and pseudo labels. In weakly supervised WSI classification, Chen \textit{et al.} \cite{chen2023rankmix} proposed a self-training framework and the concept of pseudo labeling to extract the key regions from WSIs. Liu \textit{et al.} \cite{liu2023multiple} proposed a self-paced framework to gradually improve the accuracy of pseudo labels during the training process. However, existing work mainly focuses on pseudo labels from positive slides. Instead of treating all negative patches from negative slides equally as ground truth negative labels, we intend to develop an efficient method to sample part of negative patches based on the pseudo labels from negative slides.

\myparagraph{Hard negative sample mining in WSI classification.} Hard negative sample mining was first introduced in the object detection task \cite{dalal2005histograms}, where the main idea is to repeatedly bootstrap negative samples mistakenly classified as false positives. In WSI analysis, Bejnordi \textit{et al.} \cite{bejnordi2017deep} was the first to enhance model performance on breast cancer by mining difficult negative regions from the training samples.  Furthermore, Li \textit{et al.} \cite{li2019deep} and Butke \textit{et al.} \cite{butke2021end} incorporated hard negative sample mining methods into the MIL framework to improve the performance of the WSI classification task by leveraging attention weights to identify hard negative instances in false positive bags. Unlike these hard negative mining methods that focus on training a better MIL aggregator, our method utilizes these challenging negative samples to fine-tune the encoder, leading to improved patch-level feature representation.

\begin{figure}[t]
\includegraphics[width=\textwidth]{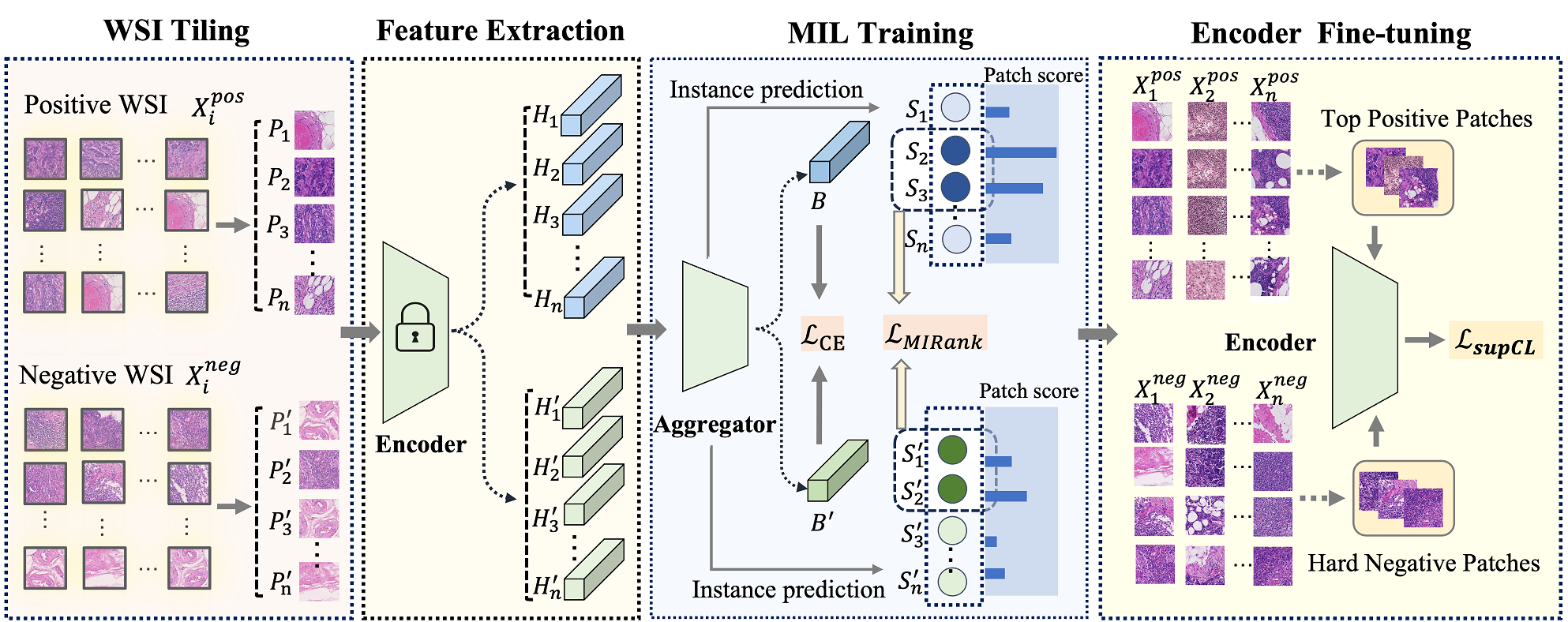}
\caption{Overview of our hard negative sample mining framework: WSIs are cut into patches. The encoder generates instance-level features which are aggregated into bag-level features and a pseudo label is assigned to each instance. The multiple instance ranking loss is employed to enhance the accuracy of the pseudo labels. Finally, negative and positive patches are selected based on enhanced pseudo labels to fine-tune the encoder and the process is repeated iteratively.}
\label{fig1}
\end{figure}

\section{Method}

\myparagraph{MIL formulation.}
In the WSI classification task, we are given a dataset $D$  consisting of a set of WSIs $ X = \{X_1, X_2, \ldots, X_N\} $ and its corresponding set of slide labels $Y = \{Y_1, Y_2, \ldots, Y_N\}$. Because WSIs are huge size images, each WSI is cut into non-overlapping smaller patches $ \{ x_{i,1}, x_{i,2} \ldots x_{i, n_i}\} $ where $n_i$ represents the number of patches obtained from $X_i$. In the setting of MIL, each WSI is considered as a bag, and all patches extracted from the WSI are considered as the instances of the bag. The bag label $Y_i \in \{0, 1\}$ and the instance labels $ \{ y_{i,1}, y_{i,2} \ldots y_{i, n_i}\} $ are unknown. 
A bag is labeled as negative only if all its instances are negative, and is labeled positive otherwise.

\myparagraph{Method overview.} The performance of a WSI classifier is tied to its instance classification performance. The main challenge is the lack of instance-level labels. Additionally, with gigapixel WSIs, the number of instances is huge (in the order of hundreds of thousands for the Camelyon16 \cite{bejnordi2017diagnostic} dataset), which negatively affects the training time. 
To improve the accuracy of instance pseudo-label prediction and training efficiency, we propose a negative sampling enhanced self-training MIL framework. \Cref{fig1} provides an overview of our proposed method. Our approach is comprised of two main components. Firstly, we incorporate a novel multiple instance ranking loss during the training of the aggregator. Subsequently, we design a more efficient strategy for negative patch sampling to fine-tune the encoder. 
We will next describe these components in more details.

\myparagraph{Multiple instance ranking loss.} 
  Let $ X_i = \{ x_{i,1}, x_{i,2} \ldots x_{i, n_i}\} $ represents a bag (WSI) and $x_{ij}$ is the $j^{th}$ instance in this bag. After extracting features using encoder $f$, each patch is projected into instance embedding $h_{ij} = f(x_{ij}) \in \mathbb{R}^{L \times 1} $. An instance classifier converts this embedding into a prediction score $s_{ij} = \phi_{ins}(h_{ij})$, $s_{ij} \in (0,1)$, where $\phi_{ins}$ are the weights of the classifier. 
  For the positive bag, the instance-level predicted scores are denoted as $ \hat{S}_i^p = \{ \hat{s}_{i,1}^p, \hat{s}_{i,2}^p \ldots \hat{s}_{i,n_i}^p\} $, and for the negative bag, the instance-level predicted scores are denoted as $ \hat{S}_i^n = \{ \hat{s}_{i,1}^n, \hat{s}_{i,2}^n \ldots \hat{s}_{i,n_i}^n\} $. The instance with the highest prediction score in the positive bag is most likely to be the true positive patch, and the instance with the highest prediction score in the negative bag is the one most similar to a positive patch but is actually negative. 
  This negative instance is considered as a hard instance.
  To push the scores of positive and negative instances far apart, 
  we propose a novel multiple instance ranking loss that aims to maximize the difference between the sum of scores of the top $K$ instances in the positive and negative bags, respectively.
The hinge-based formula of our multiple instance ranking loss is:
\begin{equation}
    \mathcal{L}_{MIRank} = \max\left(0, 1 - \frac{1}{K} \sum_{top_{K}} \hat{s}_{i,k}^p + \frac{1}{K} \sum_{top_{K}} \hat{s}_{i,k}^n \right)
\end{equation}

\myparagraph{Instance aggregator.} 
To classify the WSI, similar to \cite{li2021dual}, we first compute the bag embedding as a weighted sum of all instance embeddings. The WSI prediction is then the average of the bag classifier and the instance classifier:

\begin{equation}
  \hat{Y_{i}}  = \frac{1}{2} \left( \phi_{ins}h_m + \phi_{bag} \sum_{i} U(h_i, h_m)h_i \right)
\end{equation}
where $\phi_{ins}$ and $\phi_{bag}$ are the weights of the instance and bag classifiers, respectively. $h_m$ is the embedding of the instance with the highest score and $U(h_i, h_m)$ is the distance between $h_m$ and an instance $h_i$. Finally, the complete loss function for training the MIL aggregator is given by:

\begin{equation}
    \mathcal{L}_{MIL} = w_b * \mathcal{L}_{CE}(\hat{Y}_i, Y_i) + w_r * \mathcal{L}_{MIRank}
\end{equation}
where $\mathcal{L}_{CE}$ is the cross-entropy loss, $w_b$ and $w_r$ are the weights for the cross-entropy loss and the multiple instance ranking loss, respectively.

\myparagraph{Negative sampling enhanced contrastive learning.}
After each iteration of training the MIL aggregator, we use the trained model to obtain patch-level pseudo labels to fine-tune the encoder. Fine-tuning enables the encoder to learn discriminative representations by pulling together the representations of instances sharing the same pseudo label and pushing apart the representations of instances with different pseudo labels. Let $x$ be the anchor instance, $x_s$ is an instance selected from set $\mathcal{S}_x$ with the same pseudo label as $x$, and $x_d$ is an instance selected from set $\mathcal{D}_x$ with a pseudo label different from $x$. We use supervised contrastive learning as in \cite{khosla2020supervised,liu2023multiple} to fine-tune the encoder:

\begin{equation}
\mathcal{L}_{\text{supCL}} (x) = \frac{1}{|\mathcal{S}_x|} \sum_{x_s \in \mathcal{S}_x} -\log \frac{\text{sim}(x, x_s)}{\sum_{x_s \in \mathcal{S}_x} \text{sim}(x, x_s) + \sum_{x_d \in \mathcal{D}_x} \text{sim}(x, x_d)}
\end{equation}
The similarity score \(\text{sim}(x, x')\) is defined as 
\(\exp \left( f(x) \cdot f(x') / \tau \right)\), 
where \(f\) is an encoder, and \(\tau\) is a temperature parameter. The construction of $\mathcal{S}_x$ and $\mathcal{D}_x$ is determined by the pseudo label of $x$. Let $\mathcal{X}_{pos}$ represent the bank of positive instances, and $\mathcal{X}_{neg}$ represent the bank of negative instances. The construction of $\mathcal{S}_x$ and $\mathcal{D}_x$ is as follows:
\begin{align}
    \text{If } x \in \mathcal{X}_{pos}\text{:}
    \begin{cases}
    S_x \leftarrow \mathcal{X}_{pos} \\
    D_x \leftarrow \mathcal{X}_{neg}
    \end{cases}
    ,&\;\;\;\;\;\;\;\;
    \text{If } x \in \mathcal{X}_{neg}\text{:}
    \begin{cases}
    S_x \leftarrow \mathcal{X}_{neg} \\
    D_x \leftarrow \mathcal{X}_{pos}
    \end{cases}
    \label{eq:sample_pos}
\end{align}
where $\leftarrow$ represents random sampling from the instance bank. This means that 
if $x$ is sampled from the positive instance bank $\mathcal{X}_{pos}$, then $\mathcal{S}_x$ and $\mathcal{D}_x$ are constructed by sampling from $\mathcal{X}_{pos}$ and  $\mathcal{X}_{neg}$, respectively. 
 Similarly, if $x$ is sampled from $\mathcal{X}_{neg}$, then $\mathcal{S}_x$ and $\mathcal{D}_x$ are constructed by sampling from $\mathcal{X}_{neg}$ and $\mathcal{X}_{pos}$, respectively. 
Existing methods typically construct $\mathcal{X}_{pos}$ and $\mathcal{X}_{neg}$ as follows: 
$\mathcal{X}_{pos}$ is the collection of the top $r_p$\% of positive instances above a preset threshold.
$\mathcal{X}_{neg}$ is the collection of all instances in negative slides since, by definition, negative bags only contain negative instances. 
However, this approach is time-consuming and inefficient for fine-tuning because it includes too many easy negative instances in $\mathcal{X}_{neg}$. 
Instead, we propose to use hard negative sampling, i.e., construct $\mathcal{X}_{neg}$ from the collection of negative instances with the top $r_n$\% prediction scores.
In this way, training efficiency can be significantly improved by selecting only a fraction of hard negative instances for fine-tuning.

\section{Experiment}

\myparagraph{Datasets.} 
We conduct experiments on two public datasets: Camelyon16 \cite{bejnordi2017diagnostic} and TCGA-LUAD mutation \cite{tcga2019}. Camelyon16 is designed for detecting metastases in lymph node tissue slides. It contains 270 normal slides and 129 tumor slides. 
The TCGA-LUAD mutation dataset is aimed at detecting gene mutations. We selected four genes related to treatment options: EGFR, KRAS, STK11, and TP53 \cite{liu2023multiple,coudray2018classification}. The dataset comprises 607 WSIs, and the WSI labels indicate whether the corresponding gene is expressed in the slides. 

\myparagraph{Experiments setup and evaluation metrics.}
In the WSI preprocessing stage, we cut the slides into non-overlapping patches of size 224$\times$224. For the Camelyon16 dataset, we obtained 0.25 million patches with 5$\times$ magnification. For the TCGA-LUAD mutation dataset, we got 0.52 million patches with 10$\times$ magnification. 
We utilize the pre-trained ResNet-18 encoder provided by \cite{li2021dual} to extract features for the Camelyon16 and TCGA-LUAD mutation datasets.
We evaluate the performance on WSI classification on Camelyon16 and TCGA-LUAD mutation datasets, in addition to patch-wise classification on Camelyon16 dataset. 
We report accuracy (ACC) and area under the curve (AUC) evaluation metrics.
For Camelyon16 dataset, we reported the results of the official testing set. For TCGA-LUAD mutation dataset, we conducted 5-fold cross-validation on the 607 slides, and the mean and standard deviation of performance metrics are reported. The mean results are presented in \Cref{tab:comparison}, the detailed results with standard deviation are provided in Tabel 4 in the supplementary material. 

\myparagraph{Implementation details.}
When training the MIL aggregator, we follow the settings in \cite{li2021dual}. The MIL aggregator was trained for 350 epochs. We employ Adam optimizer with a learning rate of 0.0001. For the multiple instance ranking loss, we set $K$ to 10. The weight of cross-entropy loss $w_b$ and ranking loss $w_r$ are configured to 0.5 and 0.1, respectively. 
For both the Camelyon16 and TCGA-LUAD mutation datasets, we set the parameters for sampling pseudo labels, $r_p$ and $r_n$, to 0.2 and 0.05, respectively. The fine-tuning phase was set to 25 epochs. For these hyperparameters, we experiment with different values and select the ones best performing on the validation set. All model training and testing experiments were conducted on Nvidia A5000 GPU. 

\setlength{\tabcolsep}{7pt}
\begin{table}[t]
\small
\centering
\renewcommand{\arraystretch}{1} 
\caption{Main results on Camelyon16 and TCGA-LUAD mutation datasets.}
\captionsetup{skip=5pt}
\begin{tabular}{lccccccccc}
\Xhline{2\arrayrulewidth} 
\multicolumn{1}{c}{} & \multicolumn{2}{c}{Camelyon16} & \multicolumn{4}{c}{TCGA-LUAD mutation} \\
& & & EGFR & KRAS & STK11 & TP53 & \\ 
\cline{2-7}
Method & ACC & AUC & AUC & AUC & AUC & AUC\\
\hline
Max-pooling & 0.8295 & 0.8641 & 0.6643 & 0.5746 & 0.6702 & 0.6109 \\
ABMIL\cite{ilse2018attention} & 0.8450 & 0.8653 & 0.6848 & 0.5994 & 0.6784 & 0.6520 \\
DSMIL\cite{li2021dual} & 0.8837 & 0.9095 & 0.6956 & 0.6026 & 0.6885 & 0.6344 \\
Its2CLR\cite{liu2023multiple} & 0.9070 & 0.9465 & 0.7103 & 0.6135 & 0.7111 & 0.6703 \\
Ours & \textbf{0.9302} & \textbf{0.9604} & \textbf{0.7235}  & \textbf{0.6473} & \textbf{0.7396} & \textbf{0.7071}\\
\Xhline{2\arrayrulewidth} 
\end{tabular}
\label{tab:comparison}
\end{table}

\myparagraph{Quantitative results.} \Cref{tab:comparison} shows the comparison result on Camelyon16 and TCGA-LUAD mutation datasets. For the Camelyon16 dataset, compared with the classic MIL and self-traning methods, our method achieved the best performance, with an ACC of 0.9302 and an AUC of 0.9604. Furthermore, we also observe improved instance-level prediction accuracy (See \Cref{tab:ablation_ranking} in the ablation study section). For the TCGA-LUAD mutation dataset, our method achieved the best AUC results over four genes: EGFR reached 0.7235, KRAS reached 0.6473, STK11 reached 0.7396, and TP53 reached 0.7071. The evaluation results show the effectiveness of our proposed framework in bag and instance predictions.

\begin{figure}[t]
\small
\includegraphics[width=\textwidth]{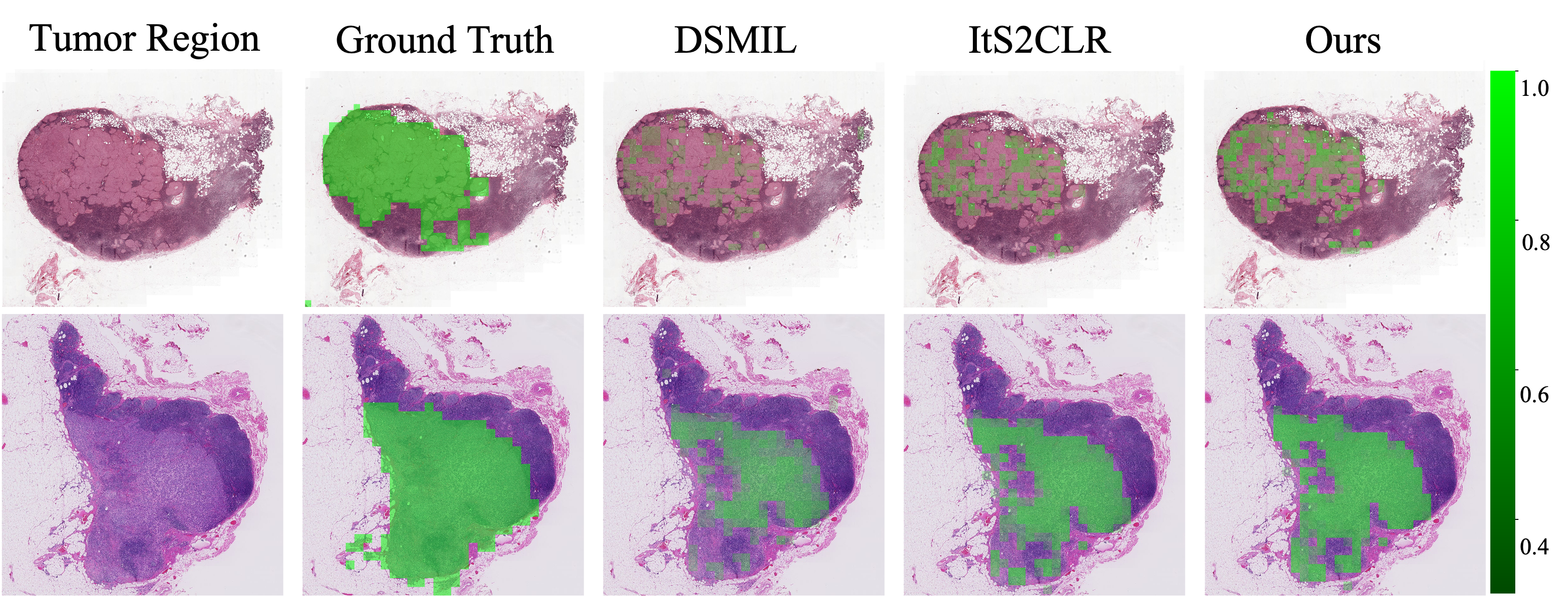}
\caption{Visualization of instance prediction probabilities on the Camelyon16 dataset. 
Patches with probabilities below 0.3 are rendered transparent.
} \label{fig3}
\end{figure}

\myparagraph{Qualitative results.} \Cref{fig3} compares the instance-level prediction in tumor-positive WSIs from the Camelyon16 dataset. Compared to the DSMIL and Its2CLR methods, the instance score maps from our method align best with the ground truth maps. The prediction score for the top negative instances gradually decreases as training progresses.
This demonstrates qualitatively that our method enhances the accuracy of instance predictions. The magnified version of \Cref{fig3} is provided in the supplementary materials (Figure 4).

\myparagraph{Ablation study: ranking loss.} To demonstrate the effectiveness of our multiple instance ranking loss (MI ranking loss), we evaluated the bag-level and instance-level performance on the Camelyon16 dataset before and after integrating our proposed loss into various methods. \Cref{tab:ablation_ranking} illustrates that incorporating our ranking loss significantly enhances prediction accuracy at both the instance and bag levels for all methods.

\begin{table}[t]
\small
\centering
\renewcommand{\arraystretch}{1} 
\caption{
{Ablation study for our proposed multiple instance ranking loss (MI ranking) with WSI and instance-wise evaluations on Camelyon16 dataset.} 
}
\captionsetup{skip=5pt}
\begin{tabular}{lccc|cc}
\Xhline{2\arrayrulewidth} 
\multicolumn{1}{c}{} & \multicolumn{3}{c}{Instances} & \multicolumn{2}{c}{WSIs}\\
\cline{2-6}
Method & ACC & AUC & AUPRC & ACC & AUC\\
\hline
DSMIL & 0.8941 & 0.9118 & 0.8876 & 0.8837 & 0.9095 \\
DSMIL + MI ranking & \textbf{0.9007} & \textbf{0.9176} & \textbf{0.8931} & \textbf{0.8914} & \textbf{0.9151} \\
\hline %
ItS2CLR & 0.9287 & 0.9478 & 0.8974 & 0.9070 & 0.9465 \\
ItS2CLR + MI ranking & \textbf{0.9291} & \textbf{0.9496} & \textbf{0.8987} & \textbf{0.9147} & \textbf{0.9483} \\
\hline %
Ours(w/o MI ranking) & 0.9341 & 0.9598 & 0.9065 & 0.9225 & 0.9583 \\
Ours & \textbf{0.9374} & \textbf{0.9619} & \textbf{0.9123} &\textbf{0.9302} & \textbf{0.9604} \\
\hline
\Xhline{2\arrayrulewidth} 
\end{tabular}
\label{tab:ablation_ranking}
\end{table}

\myparagraph{Hard negative sample size and training time.} 
We conducted experiments with negative sampling rates of 2\%, 5\%, 10\%, 20\%, and 100\%. \Cref{tab:ablation_negative} illustrates that the optimal performance is achieved at a negative sample ratio of 5\% for fine-tuning. Furthermore, we measured the training time per iteration, i.e. the time for training the aggregator, updating the pseudo labels, and fine-tuning the encoder.  
When the negative sampling rate is 100\%, the iteration time is similar to ItS2CLR since fine-tuning dominates the overall training time. However, with our proposed negative sampling strategy, the training time is significantly reduced (around $70\%$ to $80\%$ less time) compared to the ItS2CLR method and with improved performance.  
These findings demonstrate that our approach can simultaneously enhance performance and training efficiency.

\setlength{\tabcolsep}{12pt}
\begin{table}[t]
\small
\centering
\renewcommand{\arraystretch}{1} 
\caption{Ablation study on proportion of negative sample and training time.}
\captionsetup{skip=5pt}
\begin{tabular}{cccc}
\Xhline{2\arrayrulewidth} 
\multirow{2}{*}{Negative (\%)} & \multicolumn{2}{c}{Camelyon16} & \multirow{2}{*}{Training Time} \\
\cline{2-3}
& ACC & AUC & \\
\hline
2\% & 90.7 & 93.12 & 33 min / iteration \\
5\% & \textbf{93.02} & \textbf{96.04} & 39 min / iteration \\
10\% & 92.25 & 95.76 & 46 min / iteration \\
20\% & 91.47 & 95.01 & 72 min / iteration \\
100\% & 91.47 & 94.83 &240 min / iteration \\
\Xhline{2\arrayrulewidth} 
\end{tabular}
\label{tab:ablation_negative}
\end{table}

\section{Conclusion}

This work introduces a negative sampling enhanced framework designed to improve performance and training efficiency for self-training frameworks applied to WSI classification tasks. This framework consists of two components: multiple instance ranking loss and negative sampling strategy. The ranking loss enhances instance-level prediction accuracy by differentiating between positive and negative instances, and the negative instance sampling strategy selectively integrates challenging negative samples into the fine-tuning process. Extensive experiments validate the effectiveness and efficiency of our proposed framework.

\myparagraph{Acknowledgements}: This research was partially supported by the National Science Foundation (NSF) grant CCF-2144901, the National Institute of General Medical Sciences grant R01GM148970, the National Institutes of Health (NIH) grant 5R21CA258493-02, and the Stony Brook Trustees Faculty Award.

\bibliographystyle{splncs04}
\bibliography{Paper-3822}

\newpage

\title{Hard Negative Sample Mining for Whole Slide Image Classification \\ — Supplementary Material —
}

%
%

\author{Wentao Huang\thanks{Email: wenthuang@cs.stonybrook.edu.}\inst{1} \and
Xiaoling Hu\inst{2} \and
Shahira Abousamra\inst{1} \and 
Prateek Prasanna\inst{1} \and
Chao Chen\inst{1}
}
\authorrunning{W. Huang et al.}
%
\institute{Stony Brook University, Stony Brook, NY, USA \and
Harvard Medical School, Boston, MA, USA\\
}

\maketitle              

\setcounter{figure}{3}
\setcounter{table}{3}

\setlength{\tabcolsep}{3pt}
\begin{table}
\centering
\tiny
\renewcommand{\arraystretch}{1} 
\caption{Comparison result on Camelyon16 and TCGA-LUAD mutation datasets, including standard deviations.}
\captionsetup{skip=5pt}
\begin{tabular}{lccccc}
\Xhline{2\arrayrulewidth} 
\multicolumn{1}{c}{} & \multicolumn{1}{c}{Camelyon16} & \multicolumn{4}{c}{TCGA-LUAD mutation} \\
& & EGFR & KRAS & STK11 & TP53 \\ 
\cline{2-6}
Method & AUC & AUC & AUC & AUC & AUC \\
\hline
Max-pooling & 0.8641 {$\pm$ 0.0132} & 0.6643{ $\pm$ 0.0033} & 0.5746 {$\pm$ 0.0027} & 0.6702 {$\pm$ 0.0036} & 0.6109 {$\pm$ 0.0022} \\
ABMIL\cite{ilse2018attention} & 0.8652 {$\pm$ 0.0086} & 0.6848 {$\pm$ 0.0026} & 0.5994 {$\pm$ 0.0033} & 0.6784 {$\pm$ 0.0031} & 0.6520 {$\pm$ 0.0034} \\
DSMIL\cite{li2021dual} & 0.9095 {$\pm$ 0.0075} & 0.6956 {$\pm$ 0.0035} & 0.6026 {$\pm$ 0.0029} & 0.6885 {$\pm$ 0.0037} & 0.6344 {$\pm$ 0.0031} \\
Its2CLR\cite{liu2023multiple} & 0.9465 {$\pm$ 0.0023} & 0.7103 {$\pm$ 0.0030} & 0.6135 {$\pm$ 0.0025} & 0.7111 {$\pm$ 0.0032} & 0.6703 {$\pm$ 0.0024} \\
Ours & \textbf{0.9604} {$\pm$ \textbf{0.0022}} & \textbf{0.7235} {$\pm$ \textbf{0.0025}} & \textbf{0.6473} {$\pm$ \textbf{0.0034}} & \textbf{0.7396} {$\pm$ \textbf{0.0037}} & \textbf{0.7071} {$\pm$ \textbf{0.0028}} \\
\Xhline{2\arrayrulewidth} 
\end{tabular}
\label{tab:comparison-supplementary}
\end{table}


\begin{figure}
\small
\includegraphics[width=\textwidth]{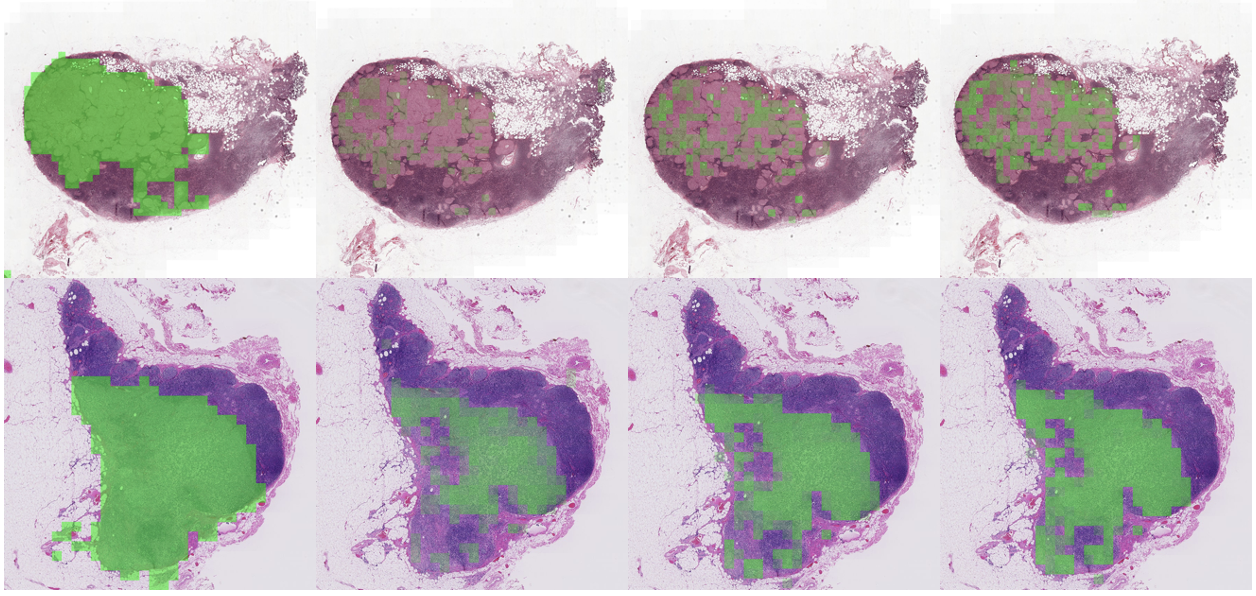}
\caption{Magnified version of Figure 3. The first column displays the instance ground truth label. The second, third, and fourth columns visualize the instance prediction probabilities generated by DSMIL, ItS2CLR, and our method, respectively. Patches with threshold probabilities below 0.3 are rendered transparent.
} \label{fig4}
\end{figure}

\begin{table}
\small
\centering
\renewcommand{\arraystretch}{1.5} 
\caption{Threshold for pseudo labeling, patches with prediction scores exceeding the threshold are marked as positive.}
\captionsetup{skip=5pt}
\begin{tabular}{lccccccc}
\Xhline{2\arrayrulewidth} 
\multicolumn{1}{c}{} & \multicolumn{1}{c}{CAMELYON16} & \multicolumn{4}{c}{TCGA-LUAD mutation} \\
& & EGFR & KRAS & STK11 & TP53 \\ 
\hline
Threshold & 0.3 & 0.1 & 0.25 & 0.1 & 0.5 \\
\Xhline{2\arrayrulewidth} 
\end{tabular}
\label{tab:pseudo}
\end{table}

%





\end{document}